\documentclass{article}
\usepackage{spconf,amsmath,graphicx}
\usepackage{cite}
\usepackage{amsmath,amssymb,amsfonts}
\usepackage{algorithmic}
\usepackage{graphicx}
\usepackage{textcomp}
\usepackage{xcolor}
\usepackage[ruled,vlined]{algorithm2e}

\usepackage{times}
\usepackage{epsfig}
\usepackage{graphicx}
\usepackage{amsmath}
\usepackage{amssymb}
\usepackage{booktabs}

\title{PT-VTON: an Image-Based Virtual Try-On Network with Progressive Pose Attention Transfer\\
}

\name{Hanhan Zhou, Tian Lan, Guru Venkataramani}
\address{George Washington University}

\begin{document}
%
\maketitle
\def\baselinestretch{0.95}
\begin{abstract}
Virtual try-on system has gained great attention due to its potential to give customers a realistic, personalized product presentation in virtualized settings. In this paper, we present PT-VTON, a novel pose-transfer-based framework for cloth transfer that enables virtual try-on with arbitrary poses.  PT-VTON can be applied to the fashion industry within minimal modification of existing systems while satisfying the overall visual fashionability and detailed fabric appearance requirements.  It enables efficient clothes transferring between model and user images with arbitrary pose and body shape. We implement a prototype of PT-VTON and demonstrate that our system can match or surpass many other approaches when facing a drastic variation of poses by preserving detailed human and fabric characteristic appearances. PT-VTON is shown to outperform alternative approaches both on machine-based quantitative metrics and qualitative results.

\end{abstract}
\section{Introduction}
By letting customers try on products from the comfort of a virtualized system, online fashion shopping has substantial benefits (e.g., time/economic efficiency and convenience) over the traditional brick-and-mortar retail model. Yet for any virtual try on system to be practical, it must have the ability to efficiently create a realistic, personalized user experience from available data - such as a limited number of photos of the clothes with or without models wearing them.
A virtual try-on system for everyone would greatly save the retailers, users, and couriers' cost and time, enhancing the user experience of customers and disrupt the way of the online fashion shopping industry. In this time of the pandemic, the need for virtual try-on has never been greater before.

Virtual try-on systems can be implemented in 2D or 3D. The 3D approach often leverages  depth cameras and other equipment to measure and rebuild the 3D model of the customer and clothes like \cite{clothcap,3Dpose,3Ddetailed}, and then renders the output with 3D models of the clothes using computer graphics or virtual reality. This approach yet would 
require more specialized and expensive equipment, and also the physical presence for body shape measurements and other manual human and machine labors for each customer and clothing, making it less practical for large-scale and everyday use.


In this paper, we focus on 2D cloth transfer, or image-based method, which generates virtual try-on results based on the users' input images with minimum cost and equipment requirements.
More precisely, given a user image, an image-based virtual try-on system would generate the user's output image wearing a designated cloth. Existing methods, either transfer the visual representation of clothes directly onto a user image \cite{17tsu_transf} - thus requiring almost identical pose of the two for appropriate results - or tend to synthesize the output image based on the image of the user and desired clothes through learning-based image systems \cite{dong2019towards,hsieh2019fit}.
However, for many of these pix2pix GAN-based works, even with a similar pose of these two, the generated image would often change or lose the physical characteristics of the users and certain texture details due to GAN's nature and the training data feeding it. 
To this end, we present PT-VTON, a novel pose-transfer-based framework for cloth transfer that generates output with natural-looking regardless of input user pose while keeping the cloth texture details and original user outlook with its corresponding body shape. 
Our multi-stage system consists of a modified state-of-the-art generative adversarial network for progressive pose attention transfer, which generates suitable model cloths with designated pose as the user, utilising a specialised training strategy that utilizes transfer learning and model images in batch training for better characteristic preservation, and two texture transfer methods. 


\begin{figure*}[h]
\begin{center}
\includegraphics[scale=0.55]{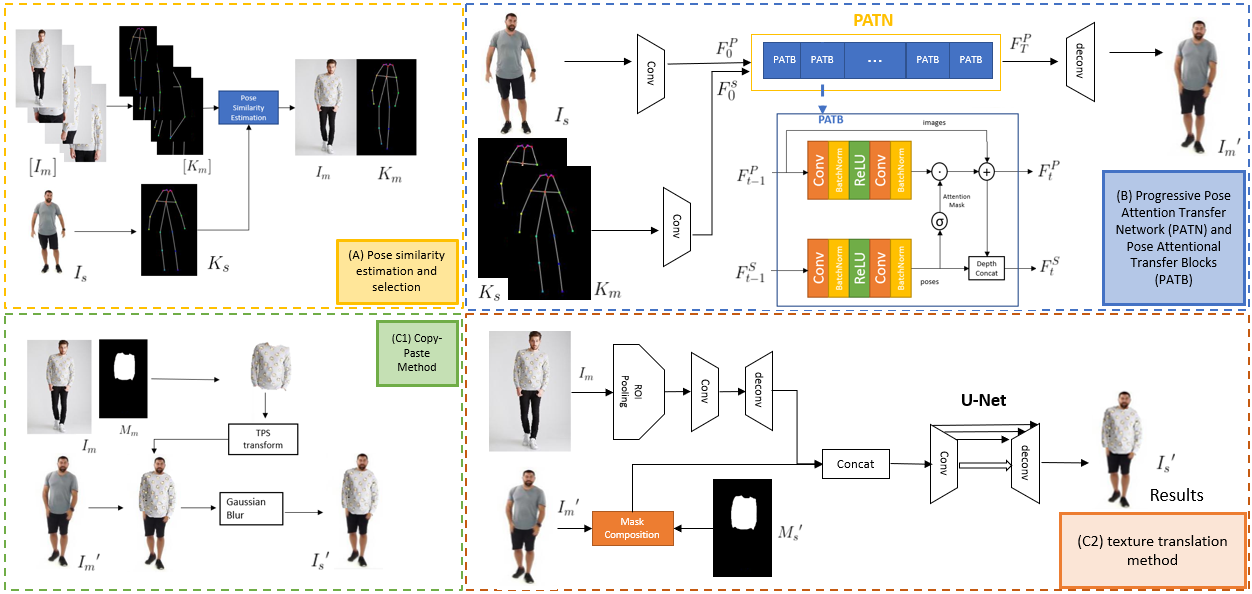}
\end{center}
   \caption{Overview of the PT-VTON framework, containing (A) Pose similarity estimation and selection, (B) A progressive pose attention transfer network and (C) two methods for texture transfer}
\label{fig:short}
\end{figure*}

PT-VTON offers a practical solution for virtual try on, since in the fashion industry, retailers are the ones to prepare such a virtual try-on system, which also possibly possess many images of models wearing different clothes in different poses, while user-provided images are often limited. Through pose transfer and our algorithm design based on transfer learning, PT-VTON is able to utilize such model images as training data in a scenario fashion industry delivering online virtual try-on services for their customers. 
In reality, customers would like to try on new clothes with different poses and see how they look in it, potentially many different poses if possible, thus a natural look with their face and body shape carefully considered are of the key requirements. In this work, PT-VTON addresses the importance of preserving user input with different poses and their original appearance. Our contributions can be summarized as follows. 


\begin{itemize}
\vspace{-0.07in}
    \item We present the a novel pose-transfer-based framework for cloth transfer (virtual try-on) that can be quickly applied to the real fashion industry within the minimal modification of existing systems.
    \vspace{-0.07in}
    \item Our approach bypasses the need for model reconstruction or any form of physical user measurements except for user images uploading.
    \vspace{-0.07in}
    \item We design a training strategy that utilizes transfer learning and model images in batch training for characteristic preservation.
    \vspace{-0.07in}
    \item We propose a multi-stage system  that accepts user image input of any pose for cloth transfer, including generating a side-view or rear-view, which many solutions do not support, paving a potential path to an image-based 360-degree virtual try-on system.
\end{itemize}

\section{Methodology}
    \vspace{-0.09in}
\subsection{Task Definition and Framework Overview}

Here we define the task as: given an image $I_{m}$ containing a person, namely model as the target pose, wearing the desired cloth for try-on and a user uploaded image $I_{s}$ which contains another person, namely user, for the specific cloth try-on, to generate the target image ${I_{s}'}$ composed of the same person as user in $I_{s}$ wearing the desired cloth model $I_{m}$ wearing as their original pose and body shape.

As Fig.3 illustrates, we first select the image  $I_{m}$ from a collection of the clothes with the most similar pose as $I_{s}$ with the help of pose similarity estimation, generate the image ${I_{m}'}$ composed of the same person as $I_{m}$ wearing the desired cloth but in the pose of $I_{s}$, with their corresponding keypoints $K_{s}$ and $K_{m}$, as the pose transfer stage. Then we transfer the texture from ${I_{m}'}$ to $I_{s}$, as they are in the same pose, with the help of a conditional segmentation mask module, determined by the appearance complexity we choose one from our two texture transfer methods.

We also propose a preparing stage, including a general training phase and a specialized training phase for the general pose transfer and the detailed cloth texture. As our experiment shows, specialized training could handle a high-quality texture transfer with more than enough capabilities to collect a new season in the fashion industry. We introduce three major refinements during this process for better results. To make the desired garment align in a characteristic position, we use pose transfer and thin plate spine transformation \cite{tps_spatial}. To perform the accurate cloth transfer without affecting the unrelated areas, we extract the garment from a pose-transferred model with the assistance of a conditional segmentation mask. To make the final results even better, we take advantage of a set of images where models wearing the same clothes with several different angles, though optional, we show by doing so, the results will look more natural with detailed textures and characters aligned at a corresponding position. 
\vspace{-0.07in}
\subsection{Pose Similarity Estimation and Transfer}
\vspace{-0.07in}
\subsubsection{Pose Similarity Estimation}

To properly choose an image with a similar pose, we first apply a flow-based pose similarity algorithm, with a comparison of object keypoint similarity between the selected image and user poses' corresponding body joints, the same metric used in Common Objects in Context (COCO) \cite{coco} dataset, where in Equation.1 $d_{i}$ are the Euclidean distances between user and model keypoint, $v_{i}$is the visibility flags for the user keypoint, $s*k_{i}$ is the standard deviation of this Gaussian times the keypoint constants from COCO. For each image, 18 joints (keypoints) are generated with OpenPose \cite{openpose} for the similarity estimation and later pose transfer. In the cloth transferring stage, we find that from a collection choosing an image with a similar pose as the user pose, meaning fewer for the generative model to infer, could produce overall better cloth texture details. 


\begin{equation} \label{eqn1}
OKS = \frac{\sum_{i}e^{-\frac{d_{i}^{2}}{2(s*k_{i})^{2}}}\delta(v_{i}>0) }{\sum_{i}\delta(v_{i}>0) }
\end{equation}
\begin{equation} \label{eqn2}
s^{*} = arg\underset{s}m\underset{\in}a\underset{S}x OKS(s)
\end{equation}
To find the image with the most similar pose as the user input, we calculate each of their Object Keypoint Similarity points (OKS) with the user input and choose the pair ($I_{m}$ and $I_{s}$) with the highest OKS points. 
\subsubsection{Pose Transfer Network and Training}
In essence, pose transfer is to move corresponding image blocks from original locations to new locations marked by target poses.The pose transfer module in our framework consists of a framework similar to Pose-Attentional Transfer Network (PATN) generator with 9 Pose-Attentional Transfer Blocks (PATB) as described in \cite{pose-transfer}. We make several changes to its framework and training procedure to make the pose transfer module suitable for transferring and memorizing detailed cloth textures.




Even with a state of the art design like \cite{Pumarola_2019_ICCV, dense, tsunashima2020uvirt,siarohin2018deformable} that performs excellent on pose transfer, it alone for the purpose of cloth transferring can be extremely challenging and unsatisfied, later shown in Fig.3(b), especially when the observation of the people and clothes are partially visible or the large difference between targeted and provided pose.Instead, when combing with the situation of the real fashion industry, where usually for a set of cloth more than one model would be wearing it and taking a collection of images wearing it with different poses. We show that taking advantage of the model images collected during the training phase and the pose transfer phase could generate better results on cloth features and texture details. 




We first train the PATN with 15,632 training pairs and 865 testing pairs from DeepFashion \cite{deepfashion} for general pose transfer training, with a small amount of overlap between the two, for 200 epochs and then we apply a specialized training with the collection of all possible clothes for later transferring with 100 pairs only for another 20 epochs. Where $L_{GAN}$ denotes generative adversarial loss and we add a parameter $\rho$ to make the pose transfer module suitable for transferring and memorizing detailed cloth textures by balancing the two discriminators in PATN.

The full and GAN loss function is as follows:
\begin{equation} \label{eqn3}
\mathcal{L} = argm\underset{G}in \:argm\underset{D}ax\:\alpha \mathcal{L}_{GAN} + \mathcal{L}_{combinedL1}
\end{equation}
\begin{equation} \label{eqn3}
\begin{aligned}
\mathcal{L}_{G A N}=& \mathbb{E}_{S_{t} \in \mathcal{P}_{S},\left(P_{c}, P_{t}\right) \in \mathcal{P}}\left\{\log \left[D_{A}\left(P_{c}, P_{t}\right) \cdot D_{S}\left(S_{t}, P_{t}\right)\right]\right\}+\\
& \mathbb{E}_{S_{t} \in \mathcal{P}_{S}, P_{c} \in \mathcal{P}, P_{g} \in \hat{\mathcal{P}}}\left\{\log \left[\left(1-D_{A}\left(P_{c}, P_{g}\right)\right)\right.\right.\\
&\left.\left.\cdot\left(1-D_{S}\left(S_{t}, P_{g}\right)\right)\right]\right\}
\end{aligned}
\end{equation}

Note that $D_{S}$ and $D_{A}$ denotes shape and appearance discriminator for their respective consistency and $\mathcal{P}$ is the probability distribution for real, fake images and their poses.
    \vspace{-0.09in}
\subsection{Texture Transfer}
    \vspace{-0.09in}
\subsubsection{Segmentation Assisted Texture Transfer}


In this stage, we perform the texture transfer based on those results from previous stages. We present two methods of texture transfer: linear copy-paste like method and a texture translation method, with the help of a human parsing segmentation mask from \cite{segmentation}. 
For the texture translation method, we applied a network similar to the texture translation module in SieveNet \cite{sievenet} , as to this stage we share the same input, being warped models and original user images in a sense. A quick description of the sub-module is shown as Fig.3., where $\mathcal{L}_{pgm}$ denotes the perceptual geometric matching loss.
\begin{equation} \label{eqn3}
\vspace{-0.5em}
\begin{aligned} 
\mathcal{L}_{warp} = \lambda_{1} \begin{vmatrix}
I_{gt}-I_{stn}^{0}
\end{vmatrix}
+\lambda_{2}\begin{vmatrix}
I_{gt}-I_{stn}^{1}
\end{vmatrix} +\lambda_{3}\mathcal{L}_{pgrm}
\end{aligned}
\end{equation}
Meanwhile, for many cases, also as shown from results in many previous works, when the main cloth segment are not largely covered by hair or any accessories, a simple copy-paste like transfer followed by a linear Gaussian process to soften the edges could also present a good result. 

Also, it's worth mentioning that there is a situation where the types of clothes between wearing and try-on are very different, e.g. when the user provided a photo of them wearing long sleeves while they intend a try-on of short sleeves, leaving part of the arms unknown to the framework. For situations like this, we would simply use the part of the arms from the model and color tune it to match the users' skin color from the appearance to avoid bias from the training data set. There isn't a better way for this type of inference to the best knowledge of us, and many previous works did not consider this situation or leave it for the framework inference.

\section{Experiments}
    \vspace{-0.09in}
\subsection{Experiment setup}
The dataset used for the experiments is a part of the In-shop Clothes Retrieval Benchmark \cite{deepfashion} namely DeepFashion and Multi-Pose Virtual try on dataset (MPV) from \cite{dong2019towards}. All images are cropped and resized to 256 x 192. We then randomly pick 100 groups of the same clothes for the second phase of specialized cloth training to simulate the new season clothes for virtual try-on. We implemented our work  with PyTorch, the hyper-parameter configuration is as follows:  a decaying learning rate starts at 0.002, $\lambda_{1}=\lambda_{2}=1, \lambda_{GAN} = 0.5$ and Adam \cite{adam} optimizer is used with $\beta_{1} = 0.5$ and $\beta_{2} = 0.9999$.
\begin{figure}[!htbp]
\vspace{-1.2em}
\includegraphics[scale=0.475]{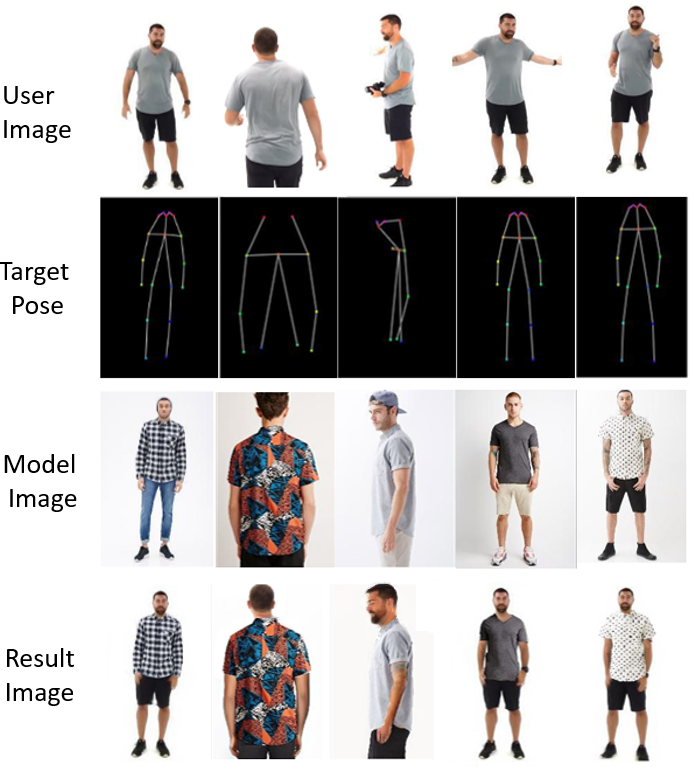}
  \caption{Qualitative results of virtual try-on with arbitrary poses and angles of view.}
\label{fig:long}
\label{fig:onecol}
\vspace{-1.25em}
\end{figure}


\begin{table}[t]
\begin{tabular}{@{}lccc@{}}
\toprule
Model                  &   IS \; \; &  \; SSIM \;   & \;  MS-SSIM \;  \\ \midrule
CP-VTON\cite{cp-vton}                & 2.66 & 0.698 & 0.746   \\
SwapNet\cite{swapnet}                & 3.04 & 0.828 &   -     \\
Sievenet\cite{sievenet}                  & 2.91 & 0.766 & 0.829   \\ \midrule
Ours(w/o LearningProc)     & 3.02 & 0.703 & 0.819   \\
Ours(Copy-Paste)       & 3.30 & 0.833 & 0.839   \\
Ours(Texture Translation) & 3.28 & 0.803 & 0.810   \\ \midrule
DataSet                & 4.18 & -     &         \\ \bottomrule

\end{tabular}
\caption{Quantitative comparison between our work and several previous works.}
\vspace{-1.25em}
\end{table}
    \vspace{-0.09in}
\subsection{Quantitative Results}
Table 1 shows a comparison of performance between our work against CP-VTON \cite{cp-vton}, SwapNet \cite{swapnet} and SieveNet \cite{sievenet} on image quality metrics benchmark, IS \cite{is} for image quality, SSIM \cite{ssim} and MS-SSIM \cite{ms-ssim} for pair-wise structural similarity. Although whether those benchmarks could define the quality of an image generated remains in debates, and some previous works like \cite{caGan}and\cite{unsupervised_pose} are based on cloth image warpping, we show improvements on those benchmarks, and for the inception score (IS) our results are closer to the dataset. Also, our framework takes an amortised time of 6 seconds to process 100 different cloth transfer requests, which features low latency for large scale industry use.
    \vspace{-0.09in}
\subsection{Qualitative Results}
\vspace{-0.09in}
The results are presented in Fig.2. We show that the overall image matches or surpasses the previously mentioned works while preserving the original user characteristics, including facial, body-looking and skin color, etc. Additionally in Fig.3 we show the comparison between our work and a coarse-to-fine framework \cite{vunet} that also swaps cloth from one to another, the results without our network modification (from the original pose transfer network) and specialised training stage as an ablation study, and some failure cases due to the mismatching of the pose keymaps.
\begin{figure}[h]
\vspace{-1.25em}
\includegraphics[scale=0.37]{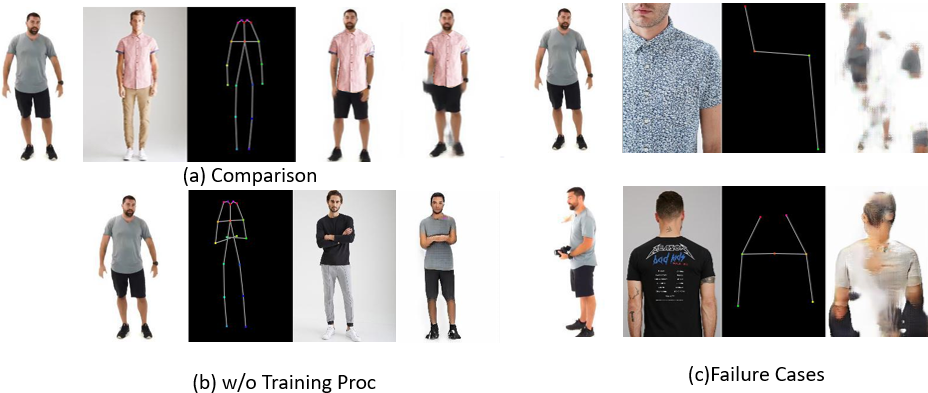}
  \caption{(a) Comparison against coarse-to-fine network. (b) Results from pose-transfer network without our modification and specialised training stage. (c) Failure cases of PT-VTON framework.}
\label{fig:long}
\label{fig:onecol}
\vspace{-1.25em}
\end{figure}

 \vspace{-0.09in}
\section{Conclusion}
    \vspace{-0.09in} 
In this paper, we propose PT-VTON, a novel framework as a solution for virtual try-on challenges with arbitrary poses. In this framework, we introduce the usage of PATN for the virtual try-on task with two-stage training procedures for better fine-grained results. Then we propose two methods for the texture transfer for different scenarios for an accurate transfer. The framework we proposed not only can be applied to industry within minor changes but is also the first virtual try-on framework that enables multiple views of angles, leaving a potential path to an image-based 360-degree surrounding virtual try-on, while preserving their original characteristic features at the same time. We show that our work qualitatively and quantitatively matches or surpasses state-of-the-art methods while providing more functionalities and possibilities.

\label{sec:ref}

{\small
\bibliographystyle{IEEEbib}
\bibliography{strings}
}

\end{document}